\documentclass{article}

\usepackage{arxiv}

\usepackage[utf8]{inputenc} 
\usepackage[T1]{fontenc}    
\usepackage{hyperref}       
\usepackage{url}            
\usepackage{booktabs}       
\usepackage{amsfonts}       
\usepackage{nicefrac}       
\usepackage{microtype}      
\usepackage{lipsum}		
\usepackage{graphicx}
\usepackage{natbib}
\usepackage{doi}
\usepackage{amsmath}

\title{Combined Optimization of Dynamics and Assimilation with End-to-End Learning on Sparse Observations}

\author{ 
Vadim Zinchenko\thanks{Corresponding author}\\
  Model-Driven Machine Learning\\
  Institute of Coastal Systems - Analysis and Modeling\\
  Helmholtz-Zentrum Hereon\\
  Germany\\
  \texttt{vadim.zinchenko@hereon.de}\\ 
   \And
 David S. Greenberg\\
  Model-Driven Machine Learning\\
  Institute of Coastal Systems - Analysis and Modeling\\
  Helmholtz-Zentrum Hereon\\
  Germany\\
  \texttt{david.greenberg@hereon.de} \\
}
\renewcommand{\shorttitle}{Combined Optimization of Dynamics and Assimilation}

\begin{document}
\maketitle
\begin{abstract}
Fitting nonlinear dynamical models to sparse and noisy observations is fundamentally challenging. Identifying dynamics requires data assimilation (DA) to estimate system states, but DA requires an accurate dynamical model. To break this deadlock we present CODA, an end-to-end optimization scheme for jointly learning dynamics and DA directly from sparse and noisy observations. A neural network is trained to carry out data accurate, efficient and parallel-in-time DA, while free parameters of the dynamical system are simultaneously optimized. We carry out end-to-end learning directly on observation data, introducing a novel learning objective that combines unrolled auto-regressive dynamics with the data- and self-consistency terms of weak-constraint 4Dvar DA. By taking into account interactions between new and existing simulation components over multiple time steps, CODA can recover initial conditions, fit unknown dynamical parameters and learn neural network-based PDE terms to match both available observations and self-consistency constraints. In addition to facilitating end-to-end learning of dynamics and providing fast, amortized, non-sequential DA, CODA provides greater robustness to model misspecification than classical DA approaches. 
\end{abstract}


\section{Introduction}
In Earth system simulators, processes we cannot represent explicitly due to incomplete understanding or prohibitive computational costs are parametrized. Compact representations of unresolved variables are dynamically updated using principled approximations or empirical models fit to observations or simulations. The free parameters of geoscientific models and their process parametrizations are most commonly adjusted through manual tuning \citep{hourdin_art_2017}, hindering reproducibility, scalability and efficient use of observation data \textit{to tune or develop parametrizations} remains rather changeable. Since real-world observations contain noise and missing values, data assimilation (DA) is required to reconstruct variable fields or `analyses' for use in parametrization learning. Essentially, we are faced with a `chicken-and-egg' problem: fitting accurate process models requires accurate analyses from data assimilation, while accurate DA requires accurate process models.

Machine learning (ML)-based parametrizations can leverage the power and flexibility of modern architectures and algorithms, including automatic differentiation and GPU support. These  outperform classical parametrizations in some cases, but have mostly been restricted to well-understood processes we can simulate cheaply enough to generate sufficient training data \citep{rasp_deep_2018, sharma_superdropnet_2024}.

However, few studies have trained ML parametrizations directly on observation data, which in practice contain both missing values and noise \citep{filoche_learning_2022, wang_four-dimensional_2024}. These mostly relied on alternating between DA and process learning, and are prone to high computational costs, estimation errors and convergence issues. These challenges are especially evident for many-parameter models (e.g. neural networks) in combination with large, spatially structured system states (e.g. 2- and 3-D variable fields in atmospheric and ocean models). Thus, new tools are required for fitting simulation parameters and training ML-based parametrizations in an end-to-end fashion on observation data.

We claim that machine learning can provide a principled and efficient framework, not only for matching fully-differentiable simulators to observations but also for process estimation directly from sparse observations. To this end, we introduce \textit{Combined Optimization of Dynamics and Assimilation} (CODA), a novel framework for solving the `chicken-and-egg' problem through gradient-based optimization of two models simultaneously: a neural network for DA and a process model with tunable parameters or an ML-based corrective term. Each model relies on the the other to produce an accurate answer, and they are trained together in an end-to-end fashion. A single loss function based on weak constraint 4Dvar incorporates both accuracy in predicting observations and internal consistency.

We apply CODA to the classical Lorenz' 96 system (L96), in both its simplified one-level version as well as the two-level scheme including unresolved fast variables. We learn the effect of unresolved scales by coupling a parametrization network to the time stepping function of slow variables. The coupled system is repeatedly self-iterated during training over multiple simulation time steps, allowing error accumulation to be evaluated and controlled. Our experiments show that our approach is effective for DA, parameter tuning and model correction tasks, matching or outperforming classical baseline methods while achieving greater robustness to model misspecification, end-to-end differentiability and full parallelization along the time axis.

\section{Methods}
In this section, we describe our approach to three important inverse problems in the geosciences: data assimilation, parameter tuning, and learning parametrizations.

\subsection{State Space Model}
We consider a numerical model or `simulator' with time-varying system state $x$ laying on evenly distributed spacial grid (e.g. air temperature, atmospheric pressure). Evolution of the system state at the next time-step $x_{t+1}$ depends on the current system state $x_t$ through time integration of a tendency function $f$.

\begin{equation}
    \label{eq:statespace}
    \frac{dx}{dt} = f(x)
\end{equation}

The system of equations is time integrated with a fixed time-step $\Delta \tau$, using a specified scheme for time integration (e.g. 4$^{\text{th}}$ order Runge-Kutta). Together, the tendency function and integration scheme define the simulator's \textit{forward operator} (also called the resolvent), which advances the system state forward in time.
\begin{equation}
    x_{t+1} = \mathcal{M}(x_t) \approx x_t + \int_0^{\Delta \tau} f(x) d\tau
\end{equation}
We assume $\mathcal{M}$ is known and fully differentiable. Starting from a system state $x_t$, we can apply $\mathcal M$ iteratively $w$ times to generate a system state sequence $x_{t:t+w}$, which we refer to as an autoregressive \textit{rollout}.

\subsection{Pseudo-observations}
In practice we typically don't observe system state variables $x_t$, but rather some other quantities $y_t$, whose distribution depends on $x_t$. In this study, we consider time-dependent observation operators $\mathcal H_t$ for which $y$ is a copy of $x$ with a (possibly random) subset of variables missing, and the observed variables corrupted by Gaussian noise.
\begin{equation}
    \label{eq:yfromx}
    y_{t} = \mathcal{H}_t(x_t) + \epsilon_t
\end{equation}
To develop and test algorithms for solving inverse problems in this system, we generate pseudo-observations by simulating system states from eq. \ref{eq:statespace} and then sampling observations from eq. \ref{eq:yfromx}.

\subsection{Data Assimilation}
In a data assimilation task, we wish to estimate $\widehat x \approx x$, given observations $y$. Critically, we do not assume access to any ground truth states $x$, either for the observations $y$ we wish to assimilate or in other scenarios.

\textbf{Hard-constraint 4Dvar} strictly incorporates observations into simulation by forcing system state $x$ to exactly match the observation $y$. It can be computationally expensive especially with a large number of observations.

\begin{equation}
    \mathcal{L}_{HC}(\widehat{x}_t) = \sum^w_{i=0} || y_{t+i} - \mathcal{H}_{t+i} \circ \mathcal{M}^{(i)}(\widehat x_t) ||^2_2
    + \left\lVert \widehat x_t - x_B \right\rVert_Q^2
\end{equation}
where $w$ and $\widehat{x}_t$ are the observation window length in time steps and estimated initial state.  $x_B$ is the background state, a prior mean arising from domain-specific assumptions or a previous DA operation with an overlapping observation window, while $Q$ is prior covariance. $\mathcal M^{(i)}$ denotes repeated autoregressive iteration of the time stepping function $i$ times, with the output of each iteration serving as the input of the next.

\textbf{Weak-constraint 4Dvar} applies the same procedure to several overlapping windows simultaneously, such that system states are consistent with each other given the dynamics $f$: that is, $\widehat x_{t_1} + \int_{t_1}^{t_2}f(x)dx= \widehat x_{t_2}$. Imposing this self-consistency constraint has been shown to significantly improve the results of data assimilation, and it forms the basis for many state-of-the-art operational methods \citep{fisher_weak-constraint_2011}. However, sequential processing of windows makes it hard to parallelize the process, and dealing with $x$ at multiple time points leads to a huge memory cost.

\textbf{Ensemble Methods} such as the ensemble Kalman smoother (EnKS), in contrast to variational data assimilation methods, leverage an ensemble of simulator-based forecasts to account for uncertainties. They incorporate real-world observations by adjusting the weights of each ensemble member based on their agreement with the data. While powerful for representing uncertainties, EnKS can be computationally expensive for high-dimensional systems and may struggle with non-linear relationships between simulator variables \citep{evensen_ensemble_2003}. 

\begin{figure}[h]
    \includegraphics[width=1.0\linewidth]{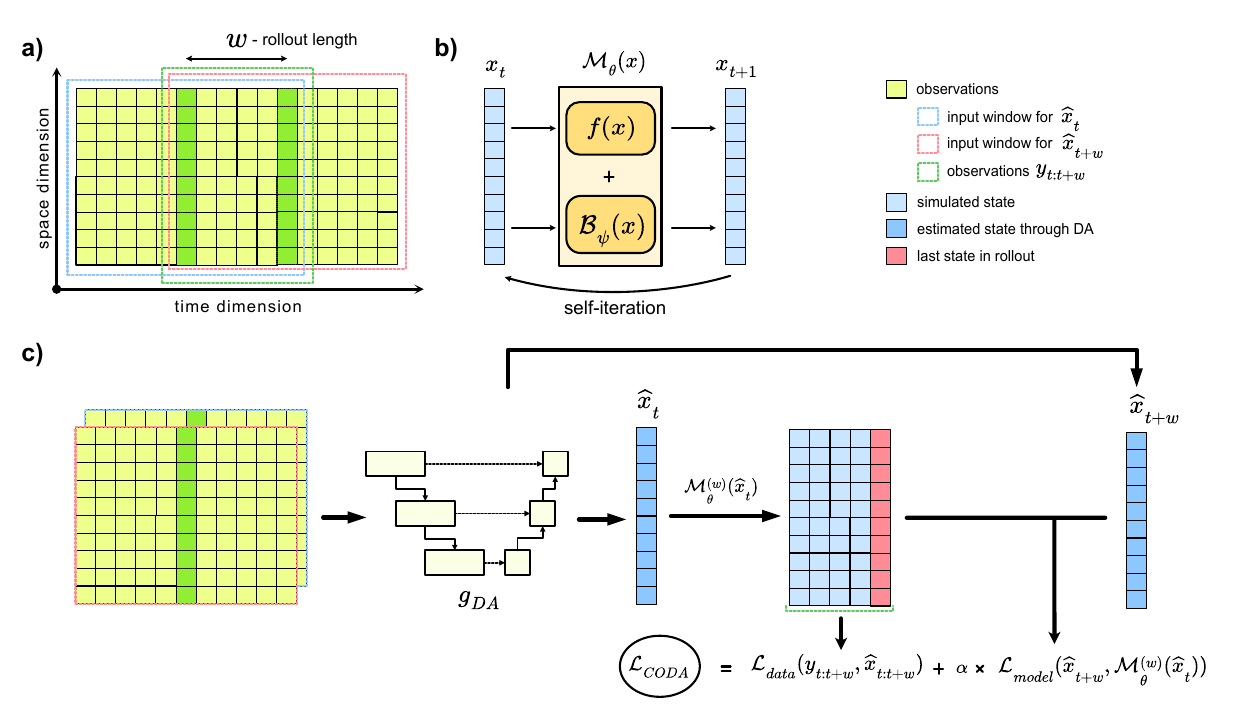}
    \caption{Learning strategy through weak constrained neural network.
    \textbf{a.} example of observations used to compute the loss $\mathcal L_\text{coda}$ for a single element of a training batch;
    \textbf{b.} autoregressive rollout through the forward operator $\mathcal{M}_{\psi}$. The learned parametrization $\mathcal{B}_{\psi}$ is an additive correction defined by a neural network;
    \textbf{c.} training step for one batch element of observations.
    }
    \label{fig:strategy}
\end{figure}

\textbf{CODA}: In order to overcome limitations of classical DA methods we introduce a novel learning strategy that substitutes 4Dvar-like optimization with a single pass through a neural network (fig. \ref{fig:strategy}). Crucially, the loss function for our ML-based DA incorporates both a data consistency term and a model consistency constraint, both of which are computed using an autoregressive rollout of simulator dynamics.

Taking as input an assimilation window of noisy and incomplete observations $y_{t-a:t+b}$, the DA network outputs an estimate $\widehat{x}_t$ of the complete, noise-free system state at time step $t$. This estimate is then used to initiate a $w$-step rollout, which is passed through the observation operators $\mathcal H_t$ to check for consistency with observations. To check for consistency with the known dynamics, the final rollout state $\mathcal M^{(w)}(\widehat{x}_t)$ is compared to $\widehat{x}_{t+w}=g_\text{DA}(y_{t+w-a:t+w+b})$, the DA network output for an assimilation window shifted $w$ steps forward in time.

\begin{eqnarray}
    \widehat{x_t} & = & g_\text{DA}(y_{t-w:t+w}) \\
    \mathcal{L}_{CODA} & = & \mathbb E_t \left [
    \underbrace{
    \sum^w_{i=0}|| y_{t+i} - \mathcal{H}_{t+i} \circ \mathcal{M}^{(i)}(\widehat x_t) ||^2_2
    }_\text{data consistency}
    +
    \overbrace{
    \alpha || \widehat{x}_{t+w} - \mathcal{M}^{(w)}(\widehat{x_t}) ||^2_2
    }^\text{model consistency}
    \right]
    \label{eq:trainingloss}
\end{eqnarray}


To apply CODA, the only requirement is that we can compute the forward operator $\mathcal M$, the observation operators $\mathcal H_t$ and their gradients.

\subsection{Parameter Tuning}
Frequently a state space model will have some parameters $\theta$ which can be adjusted, and we may be uncertain as to the correct values. Writing $\frac{dx}{dt} = f_\theta(x)$, we aim to estimate $\theta$ from a sequence of sparse, noisy observations $y$.

To extend our DA approach to parameter tuning, we optimize the same loss function $\mathcal L_{CODA}$ (eq. \ref{eq:trainingloss}) over free simulator parameters along weights of the neural network by adding a second optimizer for these parameters. Neural network and free simulator parameters are co-optimized at each training step. Neural network and free simulator parameters were jointly optimized at each training step, using the Adam optimizer without weight decay, default parameters and initial learning rate 1e-3.

\subsection{Process Correction}
We have assumed knowledge of the system dynamics $f(x)$, but in practice our understanding may be imperfect. This can occur due to our limited knowledge of relevant physical/chemical/biological processes, or because we don't have the compute budget to resolve the spatial scales of all processes needed to explain our observations $y$. We are therefore interested in learning a corrective term $B_\psi(x)$ with free parameters $\psi$, such that

\begin{equation}
    \frac{dx}{dt} \approx f(x) + B_\psi(x)
\end{equation}

In a process correction task, we know $f(x)$ but wish to estimate the corrective term $B(x)$. In this work restrict our attention to local corrections, where $B_\psi(x)$ is computed for each location of $x$ separately. This is a common scenario (turbulence closures, source/sink terms in biogeochemical simulations, etc.). We use a neural network to represent $B_\psi$, separate from the DA network used to estimate $\widehat x$ from $y$.

We employ end-to-end learning, with DA and process correction networks trained simultaneously to minimize $\mathcal L_{CODA}$. Practically this means we minimize both the mismatch between the reconstructed trajectory and the observations and the mismatch between assimilated states and model dynamics, to train both the DA network and the corrective term.

\subsection{Network Architecture}
\label{subsec:arch}
\begin{figure}
    \includegraphics[width=1.0\linewidth]{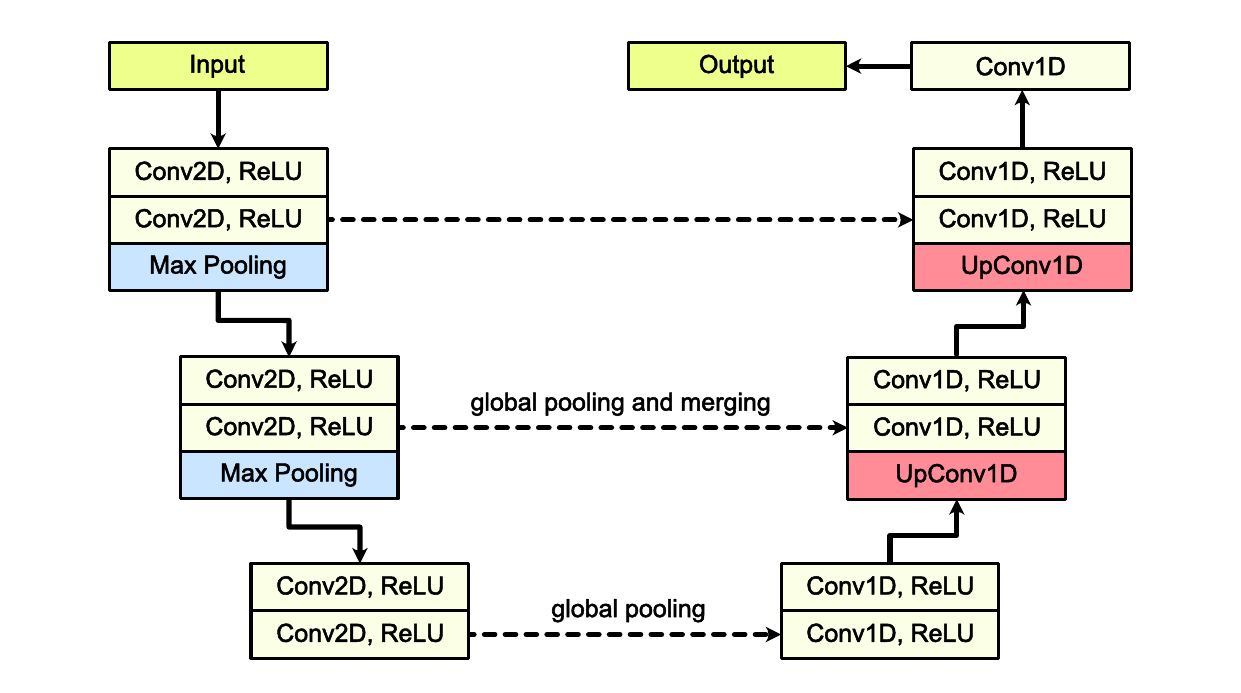}
    \caption{1.5D Unet Architecture for 1D data assimilation applications.}
    \label{fig:danet}
\end{figure}

Since in this work we are concerned with 1D systems, network inputs are observations including a time dimension and hence 2D, while outputs are 1D. We therefore propose a novel 1.5D Unet (fig. \ref{fig:danet}). The encoder has 2D convolution layers to extract important information across space and time at multiple spatio-temporal scales. The decoder uses 1D convolution layers to gather extracted information from all scales, and estimates the initial condition of the state space. Each convolution layer is followed by a batch normalization layer on both sides of the network. Like a standard Unet \citep{ronneberger_u-net_2015}, our architecture connects encoder and decoder layers blocks via skip connections, which in our case follow global max pooling operations along the time dimension. Unets allow information to pass over long distances while learning and exploiting image features on multiple spatial scales, and have been effective for segmentation \citep{ronneberger_u-net_2015}, image manipulation \citep{isola_image--image_2018}, PDE integration \citep{gupta_towards_2023, lippe_pde-refiner_2023}.

\section{Experiments}
In this section, we show the application of a novel neural data assimilation framework for three important problems in geoscience: data assimilation, parameter tuning, and learning unresolved processes. We show our results applying machine learning driven data assimilation on Lorenz'96 simulator \citep{lorenz_edward_n_predictability_1996}.

\subsection{1-level Lorenz'96 Simulator}

\begin{equation}
    \frac{dx_k}{dt} = -x_{k-1} (x_{k-2} - x_{k+1}) - x_k + F
    \label{eq:l96}
\end{equation}
a constant forcing parameter parameter $F = 8$ and boundary conditions are periodic $x_{K+1} = x_1$, $x_0 = x_K$, and $x_{-1} = x_{K-1}$. We use the time-step of $dt = 0.01$, the dimension of the system state $K=40$.

\subsection{Data Assimilation}
\begin{figure}
    \includegraphics[width=1.0\linewidth]{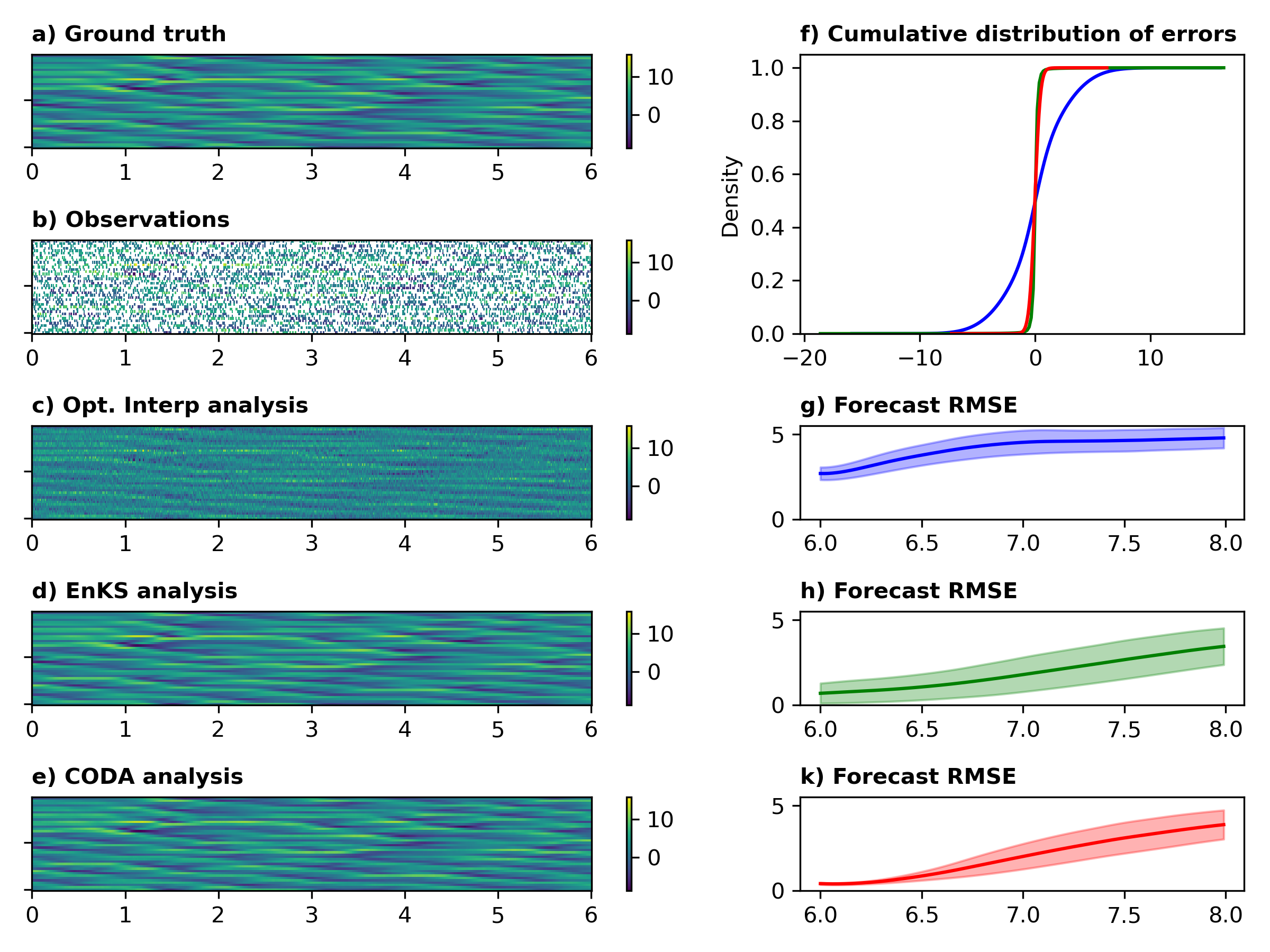}
    \caption{Comparison of CODA with baseline DA approaches.
    \textbf{a.} ground truth simulation generated using one level L96 simulator with $F=8$ and $dt=0.01$;
    \textbf{b.} observations generated from ground truth simulation applying Gaussian noise with standard deviation of $1.0$;
    \textbf{c.-e} analysis field using Optimal Interpolation (OI), Ensemble Kalman Smoother (EnKS) and CODA.;
    \textbf{f.} cumulative distribution of errors with respect to ground truth simulation (blue - OI, green - EnKS, red - CODA);
    \textbf{g.-k.} Forecast root-mean-square error started with estimated initial condition by OI, EnKS and CODA. Forecast RMSE mean (line) and standard deviation (shading) are computed for 1000 independent sets of observations generated from different initial conditions.
    }
    \label{fig:daexample}
\end{figure}

We trained a neural network to carry out data assimilation on L96 pseudo-observations with 75\% missing observations and Gaussian noise with $\sigma=1$ (fig. \ref{fig:daexample}a-b). This DA network, for which we introduce a novel 1.5D Unet architecture (fig. \ref{fig:danet}, details in \ref{subsec:arch}) took observations as input and estimated the complete denoised system state at one time point. We trained it on a sum of data-mismatch and self-consistency loss terms, both of which are computed by initialization simulations with assimilated states (eq. \ref{eq:trainingloss}). We initially evaluated DA networks using simulations and pseudo-observations outside their training sets, but with the same noise levels and observation patterns.

By shifting the DA network's input window across the observations, the full system state trajectory or `analysis' was reconstructed (fig. \ref{fig:daexample}e). Comparing the analysis to ground truth simulations (which were not available to the learning algorithm or DA network), we evaluated the accuracy of recovered system states for our DA networks as well as several alternative methods (fig. \ref{fig:daexample}c-f). Overall, we found that DA networks, which parallelize computation on the time axis, were comparable in accuracy to EnKS, despite the latter operating sequentially over time. Both our approach and EnKS strongly outperform the optimal interpolation baseline.

We further compared DA methods in a forecasting task, in which a simulation initialized by DA extended beyond the last available observation (fig. \ref{fig:daexample}g-k). Simulations initialized by the DA network closely predicted ground truth L96 values for short forecast horizons, but as expected for a chaotic system, after $5$ time units beyond the final observation were no more accurate than a random sample from the stationary distribution. In contrast, optimal interpolation-based forecasts exhibited a much higher RMSE of $2.6$ immediately after the final observation, and were equivalent to random guessing after only $2$ time units. EnKS-based forecasts exhibited similar RMSEs to the neural network, but with greater variability. Overall, these results indicate that our DA networks are highly competitive with existing DA methods in terms of cost and accuracy, while providing parallelization over time and full differentiability.

\begin{figure}[h]
    \centering
    \includegraphics[width=1.0\linewidth]{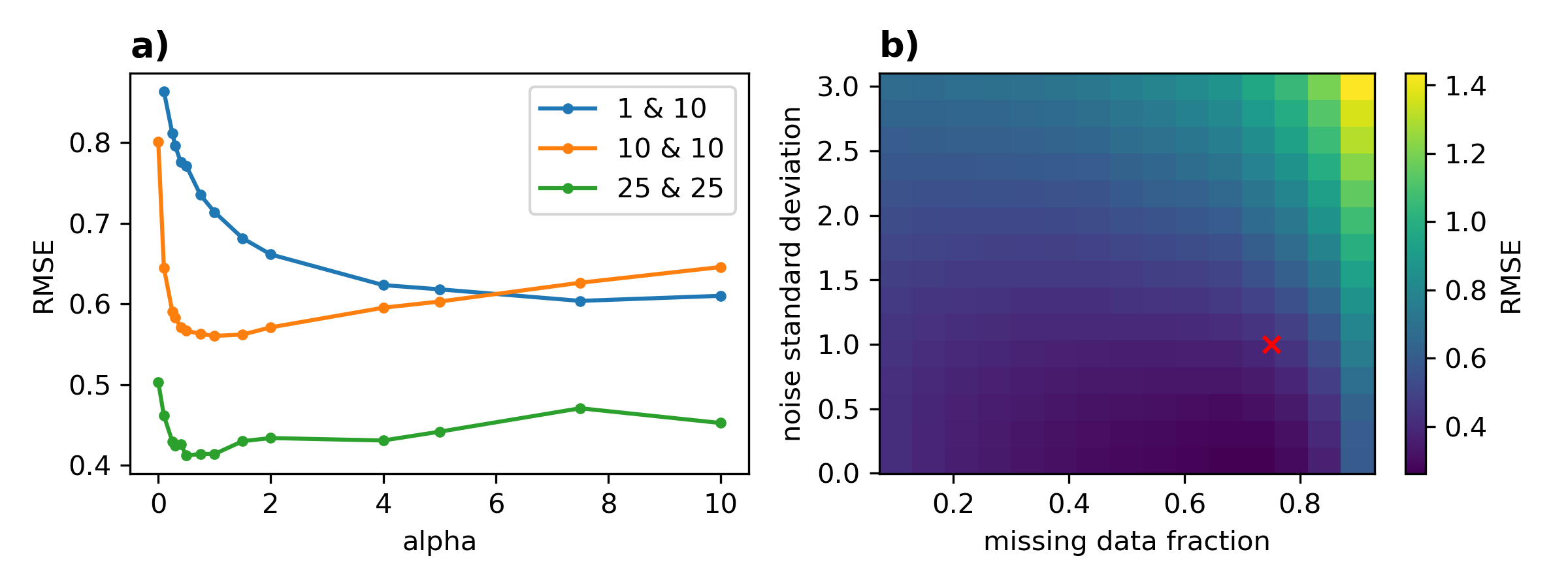}
    \caption{ Root mean square error of CODA analysis,
    \textbf{a.} different training setups including strong constraint ($\alpha=0$);
    \textbf{b.} generalization of best DA network trained on data with $75\%$ missing data and  noise standard deviation $1.0$;
    }
    \label{fig:datest}
\end{figure}

We also examined how the performance of DA networks was affected by the choice of window size for input observations and the length of simulation rollout used to compute the loss function, as well as the regularization strength $\alpha$ in eq. \ref{eq:trainingloss}. Intuitively, we would expect that longer observation windows would improve DA accuracy, but would require larger networks and possibly more training data. We would also expect use of longer rollouts to result in better-regularized and more accurate DA, but for sufficiently long rollouts the chaotic dynamics of L96 would result in excessively steep loss landscapes that would interfere with gradient-based learning. In practice, we observed a uniform improvement in agreement with ground truth system states as the lengths of both the observation window and the rollout increased (fig. \ref{fig:datest}a). For all window and rollout lengths, we found that regularization through a self-consistency term improved results, though longer windows and rollouts required less regularization, and were less sensitive to the value of $\alpha$. These results show that a DA network can effectively use extended observation windows and long rollouts, and benefit from weak constraint-like regularization terms.

Finally, we tested how DA networks trained on 75\% missing observations and noise level $\sigma=1$ would perform with other observation operators. We found that when comparing DA networks' analyses to ground truth system states, RMSE varied smoothly with the fraction of missing data and observation noise, and RMSE was less than half the observation noise, even when including unobserved values, for all settings tested up to 95\% missing values and $\sigma=3.0$. These results show that trained DA networks are robust to changes in the observation process. Based on these findings, we fixed the window size and rollout length for all subsequent experiments involving DA networks to 25 time steps.

\subsection{Parameter Tuning}
We next extended our deep learning-based DA approach to incorporate parameter tuning from observation data, a more challenging inverse problem than DA alone. Here our task is to identify values for parameters of a mechanistic dynamical model, such that noisy and incomplete observations become consistent with a simulation starting from unknown initial conditions. A strong advantage to extending our deep learning-based DA approach to this task is that the parallel-in-time nature and differentiability of our system allow for an end to end-learning approach. We pursued this possibility by optimizing jointly over the simulation's free parameters and many of the parameters of the DA network, using the same objective function as previously (eq. \ref{eq:trainingloss}).

\begin{figure}[h]
    \includegraphics[width=1.0\linewidth]{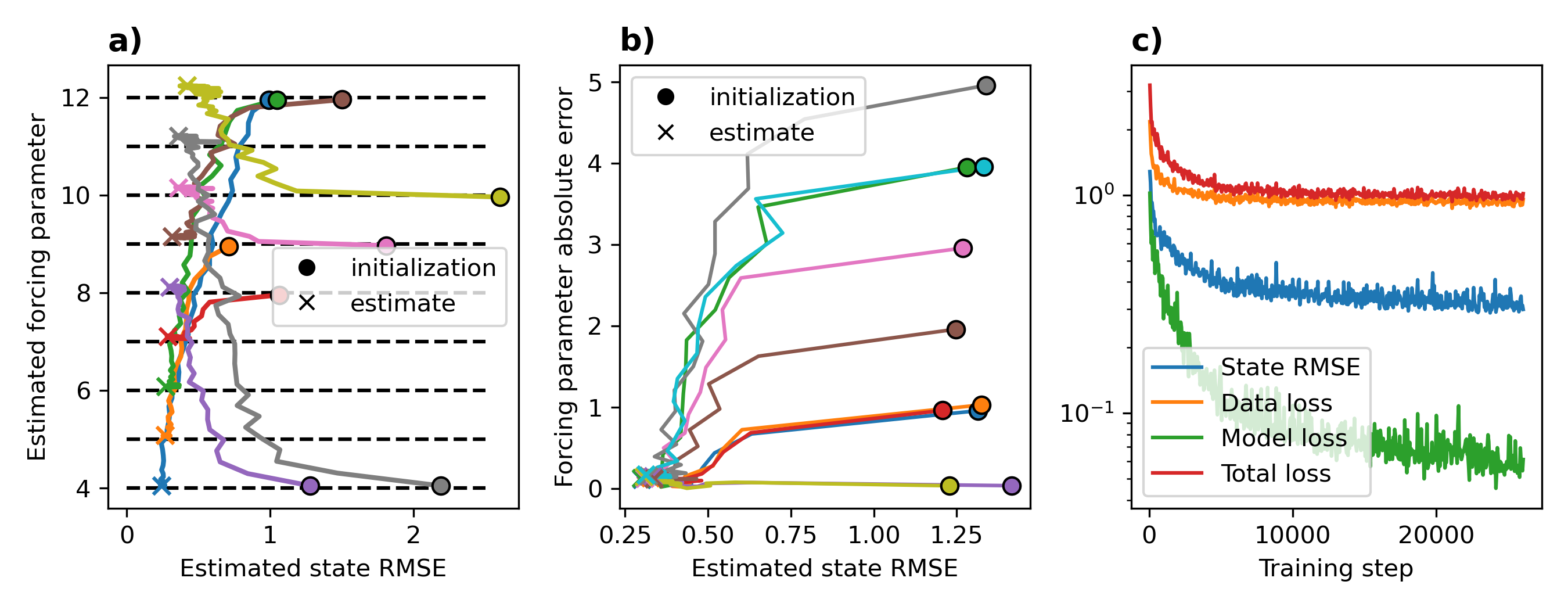}
    \caption{Convergence of CODA-based parameter tuning task along training data assimilation network. 
    \textbf{a.} 1 random initialization per task with various f;
    \textbf{b.}10 random initializations of parameter tuning task for $f=8$;
    \textbf{c.} root mean squared error of estimated initial state and loss terms: data mismatch, model error and total loss.
    }
    \label{fig:pttest}
\end{figure}

As a demonstration, we trained DA network parameters while simultaneously optimizing over the 1-level L96 parameter $F$ (eq. \ref{eq:l96}). $F$ strongly influences the behavior of the system, with higher values leading to less predictable and more chaotic dynamics (fig. \ref{fig:pttest}a). We carried out experiments in which L96 simulations were generated using $F$ values ranging from 4 to 12, and $F$ and DA network parameters were jointly optimized from pseudo-observations (75 \% missing values, noise $\sigma=1$). Over successive gradient steps, we observed what happened to the performance of DA, the loss terms and the accuracy of F (fig. \ref{fig:pttest}c). In each experiment, the randomly initialized estimate of $F$ converged over gradient steps close to the true value, while the RMSE between the analysis and ground truth decreased. To determine whether estimation of $F$ from noisy and incomplete observations was robust with respect to random initialization of DA network parameters and $F$ itself, we measured convergence of $F$ and the analysis RMSE for multiple random seeds with the true $F=8$ (fig. \ref{fig:pttest}b). Analysis RMSE and $F$ converged to similar values for all runs, indicating that our end-to-end learning approach can consistently recover simulation parameters from sparse and noisy observations, without access to ground truth system states at any point in the learning process.

\subsection{Learning Corrective Terms}
In some cases, model uncertainty cannot be represented using a small number of unknown mechanistic parameters, but instead involves uncertainty about the fundamental nature and basic mathematical form of certain aspects of the system state's evolution. We are then faced with the more challenging task of learning a corrective term for a dynamical system directly from observations. This is essential when we lack a first-principles quantitative model, such as when modeling certain biological processes or ecosystems. Corrective terms also appear frequently in atmospheric and ocean models as representations for well-understood physical processes that cannot be resolved at computational feasible resolutions. In both cases a hybrid modeling approach is commonly employed: a known but incomplete process model from physics or another domain must be supplemented with a correction representing unknown or unresolved effects. Deep neural networks can learn and compute these corrections, and their outputs can be used to correct the system's time stepping function, or to correct the tendency function inside a time stepping scheme (e.g. RK4).

As for parameter tuning, we pursue an end-to-end approach in which the DA network and correction network are simultaneously trained. We demonstrate this approach in the two-level Lorenz' 96 system formulation, whose periodic 1-D system state contains 36 coarse-scale variables and 360 fine-scale variables. Coarse- and fine-scale variables exhibit wave-like patterns propagating clockwise, and interact with each other through coupling terms:

\begin{eqnarray}
    \frac{dx_k}{dt} &=& -x_{k-1} (x_{k-2} - x_{k+1}) - x_k + F - hc \bar{z}_k
    \label{eq:l96slow}
    \\
    \frac{1}{c} \frac{dz_{j, k}}{dt} &=& -bz_{j+1, k} (z_{j+2, k} - z_{j-1, k}) - z_{j, k} + \frac{h}{J} x_k
    \label{eq:l96fast}
\end{eqnarray}

We use the time-step of $dt = 0.01$, dimensions of system states $K=36$ and $J=10$, forcing and coupling parameters $F = b = c = 10$, $h = 1$ and $\bar{z}_k = \frac{1}{J}\sum_j z_{j, k}$. We time-integrate the discretized ODE using explicit RK4 integration. We note that when $hc=0$ the evolution of the coarse variables $x$ matches the 1-level L96 system used previously.

Our task in this system is to learn a corrective term to parametrize unresolved fine-scale variables, such that replacing the coupling term $hc \bar{z}_k$ in eq. \ref{eq:l96slow} with a deep learning-based corrective term $\mathcal{B}_{\psi}$ yields coarse variables $\tilde x$ with the same dynamics as the coarse variables of the full 2-level system:

\begin{equation}
    \label{eq:l96corrected}
    \frac{d \tilde x_k}{dt} = -\tilde x_{k-1} (\tilde x_{k-2} - \tilde x_{k+1}) - \tilde x_k + F + \mathcal{B}_{\psi}(\tilde x_k)
\end{equation}

We aim to choose neural network parameters $\psi$ such that $\tilde x_t \approx x_t, \forall t$. For training, we have access to noisy and incomplete pseudo-observations $y$ (75\% missing, $\sigma=1$) of the 36 coarse-scale variables $x$, and no observations whatsoever of the fine-scale variables $z$.

\begin{figure}
    \includegraphics[width=1.0\linewidth]{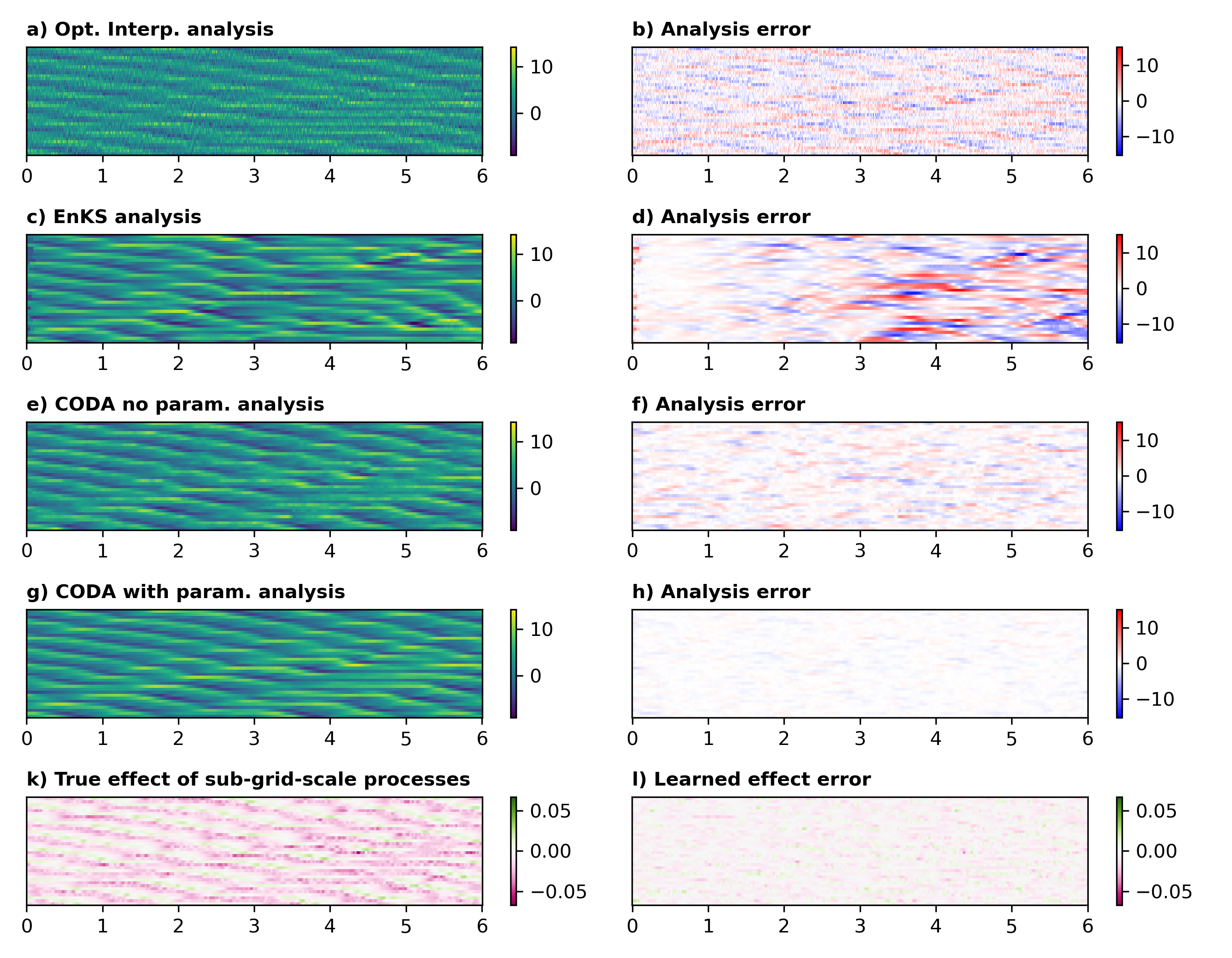}
    \caption{      
    Analysis fields obtained from sparse and noisy observations using OI \textbf{(a)}, EnKS \textbf{(c)}, CODA with an imperfect forward operator $\mathcal M$ \textbf{(e)} and CODA combined with a neural network parametrization \textbf{(g)}. Analysis errors are computed with respect to the ground truth simulation, which was never available to the DA network  \textbf{(b, d, f, h)}. True effect of sub-grid-scale parametrization \textbf{(k)} and error of learned representation by neural network \textbf{(l)}.
    }
    \label{fig:daparam}
\end{figure}

As a first step towards learning a corrective term, we trained the DA network to produce analyses for $x$ using observations $y$. To compute the loss in eq. \ref{eq:trainingloss}, we used the 1-level tendency function (eq. \ref{eq:l96}) containing neither the coupling term $hc \bar{z}_k$ nor its learned approximation $\mathcal B_\psi(x_k)$. We refer to this scenario, in which pseudo-observations are derived from the dynamics of the full state space ($x, z$, eq.\ref{eq:l96slow}-\ref{eq:l96fast}) but the simulation rollouts used to compute the loss in eq. \ref{eq:trainingloss} omit some terms and state space variables, as \textit{learning with a truncated model}. When training a DA network on the truncated model, we recovered analyses that captured the wavelike structures but exhibited some incorrect wave positions and shapes, as well as some high frequency speckle noise (fig. \ref{fig:daparam}e, g). In contrast, classical DA algorithms such as OI and EnKS were less able to cope with model misspecification, presumably since there was no learning process to compensate for model errors (fig. \ref{fig:daparam}a-d). We further observed that EnKS, the only sequential algorithm tested, was unique in exhibiting error accumulation over time. These results show that our learning approach can produce successful DA results from observations alone, even when the model exhibits systemic deficiencies.

We next used DA network parameters that had converged after learning with a truncated model to initialize further training of the DA network jointly with a second network computing the local corrective term. The RMSE between the DA network's analysis and ground truth improved during joint training with the corrective term, and all loss terms decreased (fig. \ref{fig:daparam} g, h). The effect of the learned parametrization $\mathcal B_\psi$ over complete RK4-integrated time steps was similar to the effect of coupling terms (fig. \ref{fig:daparam} k, l), though the learned correction was smoother in time and space since all fluctuations in fine-scale variables could not be predicted from a single slow variable.

\begin{figure}
    \includegraphics[width=1.0\linewidth]{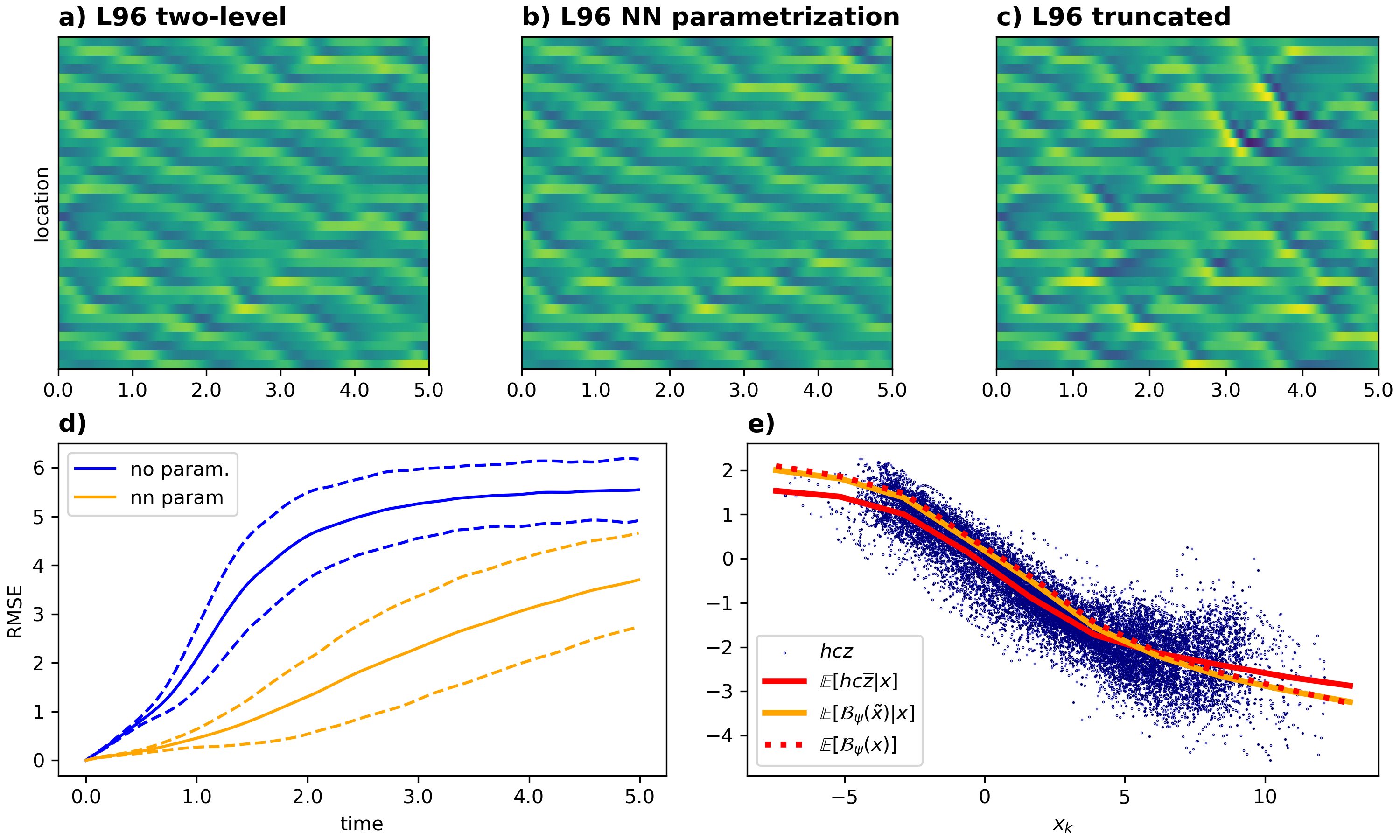}
    \caption{ 
    \textbf{a)} trajectory of slow variables in two-level L96 simulation. 
    \textbf{b)} trajectory of slow variables in hybrid L96 simulator using truncated system and CODA-trained neural network parametrization, with the same initial conditions as `a.'
    \textbf{c)} trajectory of truncated L96 system, with the same initial conditions as `a-b.'
    \textbf{d)} RMSE over time when comparing a truncated L96 system to the slow variables of a two-level L96 simulation with and without the CODA-trained parametrization, averaged over 1,000 initializations.   
    \textbf{e)} parametrization input output-function learned from sparse observations using CODA. Blue dots show true feedback from fine-scale variables, solid red line shows average true effect of fine-scale variables as a function of the local coarse-scale variable, yellow line shows parametrization output computed using analysis state $\widehat x$ as a function of the true coarse state $x$, and dashed red line shows parametrization output as a function of true coarse state.
    }
    \label{fig:parameval}
\end{figure}

After training, we evaluated the learned parameterization in free-running simulations by time integrating eq. \ref{eq:l96corrected}, and compared results to 2-level L96 simulations with the same initial conditions for coarse variables $x$. The corrected model produced rollouts that closely resembled the coarse variables of a two-level scheme, while rollouts produced by the truncated (1-level) model exhibited different trajectories and wave patterns (fig. \ref{fig:parameval}a-c). The learned correction reduced RMSE vs. ground truth by several standard deviations (fig. \ref{fig:parameval}d), and while the truncated model was no closer to ground truth than a randomly selected time point after 4 time units, the corrected rollout was more significantly accurate than random guessing for $>5$ time units. Comparison of the corrective term to the actual coupling term showed that the learned correction closely predicted the average coupling term as a function of $x$ (fig. \ref{fig:parameval}e).

Interestingly, the learned correction overestimated the coupling term for the most negative $x$ and underestimated it for the most positive $x$; this would have the effect of guiding longer corrected rollouts towards back to the attractor for $x$ and avoiding divergence to infinity, and may arise from to the use of rollouts in training (eq. \ref{eq:trainingloss}). Finally, to determine whether errors in data assimilation affected the results of the correction term, we compared the correction $\mathcal B_\psi(x)$ to $\mathbb E[\mathcal B_\psi(\tilde x) |x]$, where $\tilde x$ is the assimilated state and $x$ the ground truth state. This revealed essentially no average difference between the corrective term evaluated on the ground truth $x$ or the analysis states $\tilde x$, suggesting that DA errors do not play a major role in the differences observed between the corrective term and the coupling term of the full 2-level L96 system.

Overall, these results show that end-to-end learning can effectively parametrize unresolved spatial scales with access only to sparse, noisy observations. The resulting parametrizations improve the accuracy of both data assimilation and free-running simulations with the corrected model.

\section{Discussion}
We trained data assimilation networks directly on sparse and noisy observations, without requiring access to a ground truth reference arising from previous DA or simulations. Our approach is fully parallel in time but achieved better or comparable accuracy to sequential DA. It was also more robust to model misspecification, as learning can adapt to mismatches between the simulator and observation data. A critical factor in the success of this approach was the data-independent self-consistency loss term, inspired from weak constraint 4DVar methods.

\subsection{Related Work}
Most previous studies developing ML-based DA focused on supervised learning. Neural networks have been trained to map directly from pseudo-observations to simulated system states \citep{cintra_tracking_2016,frerix_variational_2021,penny_integrating_2022}, or from real observations to analysis datasets produced by classical DA algorithms \citep{de_almeida_neural_2022}. While simple and effective, this has significant drawbacks: pseudo-observations inherit uncorrected model errors, while analyses are expensive and limited to about half a century of global coverage. Since its training targets consist of a fixed sequence of system states, supervised learning is poorly suited to tuning or correcting the underlying simulator. A notable exception is 4DVarNet \citep{fablet_learning_2021}, a parallel-in-time supervised method that updates system states using gradients of a learned energy function, with no physical model.

Our three-step approach, in which (1) a neural network maps observations onto initial conditions to be (2) time integrated with (partially) known dynamics before (3) computing an unsupervised variational loss on observations, is shared by another recent study \citep{filoche_learning_2022}. This work applied a hard-constraint 4Dvar objective ($\alpha=0$) with no background term to partially observed 8-step L96 sequences to train a DA network. A closely related approach \citep{wang_four-dimensional_2024} trained a novel architecture to compute initial states, which were then plugged into known dynamics to calculate a variational loss (on classical DA-based analyses instead of observations). Our results go further in several ways. Most critically, we introduce a self-consistency constraint (second term in eq. \ref{eq:trainingloss}) that significantly improves DA accuracy (fig. \ref{fig:datest}). Furthermore, while both previous studies assumed a perfect forward operator, we successfully demonstrated DA for misspecified models, parameter tuning and parametrization learning.

Few principled methods exist for tuning geoscientific models, and these are mostly based on ensemble Kalman filtering \citep{ruckstuhl_parameter_2018}. Alternatively, simulation-basfed inference approaches can map observations directly onto parameters \citep{qu_joint_2024}, or estimate a posterior distribution on parameters \citep{cranmer_frontier_2020}, but require many simulations for training and cannot scale to the large systems encountered in the geosciences.

Unsupervised model correction is even more challenging, since accurate DA and an accurate process model require each other. Several studies trained model corrections on observations indirectly, by applying supervised learning to analysis datasets produced by classical strong-constraint 4Dvar \citep{farchi_using_2021}, weak-constraint 4Dvar \citep{bocquet_data_2019,bocquet_bayesian_2020} or gradient-free ensemble methods \citep{brajard_combining_2020-1,brajard_combining_2020}. Without end-to-end minimization of a consistent objective function, these techniques can be hindered by slow or inconsistent convergence, may require many rounds of alternating DA and learning, and exhibit the inefficiency of classical time-sequential DA.

\subsection{Future Outlook}
We anticipate a number of possible extensions to this work. While we did explore various values for some hyper-parameters, it is likely that significant performance gains can be made for DA, tuning and model correction with better hyper-parameters, as well as separate optimization settings for the simulator, model correction networks and DA networks. Other architectures could also prove more effective than our 1.5D Unets.

For model tuning and correction, an advantage of our unsupervised learning strategy of joint end-to-end optimization is that it trains the DA network specifically for the observations in our training dataset, as opposed to wasting effort on learning DA for any possible input data. However, it might also be desirable to produce an \textit{amortized} DA network that produces correct answers for a wider range of initial conditions or simulator parameters.

While DA was less accurate for 1-step rollouts, the fact these could be successfully used for DA with weak-constraint regularization  was encouraging, particularly in the context of large Earth system models for which longer rollouts may not fit in memory. In this case, it could make sense to choose the assimilation window and network architecture so that the memory and computation costs of inference with a DA network become comparable to 1-step rollout. This would mean that neither the DA nor simulation operations would be bottlenecked waiting for the other.

This approach aims to provide a next step towards developing hybrid simulators for the Earth system. Learned processes can be coupled back to oceanic and atmospheric simulators providing better representation of poorly understood processes (interactions between species in an ecosystem, chemical reactions, turbulent processes, and etc.). This could improve both process representations and forecast accuracy. In the longer term, as fully differentiable simulators of the atmosphere and ocean develop into full-featured status and gain widespread use, the algorithms developed here could be used to tune all their parameters and parametrizations to sparse and noisy observation data.

\section*{Acknowledgments}
V. Zinchenko is supported by Coastal Futures project, and D.S. Greenberg by Helmholtz AI. We thank Tobias Schanz for assistance implementing hydra-based configuration management.

\bibliographystyle{unsrtnat}

\begin{thebibliography}{25}
\providecommand{\natexlab}[1]{#1}
\providecommand{\url}[1]{\texttt{#1}}
\expandafter\ifx\csname urlstyle\endcsname\relax
  \providecommand{\doi}[1]{doi: #1}\else
  \providecommand{\doi}{doi: \begingroup \urlstyle{rm}\Url}\fi

\bibitem[Hourdin et~al.(2017)Hourdin, Mauritsen, Gettelman, Golaz, Balaji, Duan, Folini, Ji, Klocke, Qian, Rauser, Rio, Tomassini, Watanabe, and Williamson]{hourdin_art_2017}
Frédéric Hourdin, Thorsten Mauritsen, Andrew Gettelman, Jean-Christophe Golaz, Venkatramani Balaji, Qingyun Duan, Doris Folini, Duoying Ji, Daniel Klocke, Yun Qian, Florian Rauser, Catherine Rio, Lorenzo Tomassini, Masahiro Watanabe, and Daniel Williamson.
\newblock The {Art} and {Science} of {Climate} {Model} {Tuning}.
\newblock \emph{Bulletin of the American Meteorological Society}, 98\penalty0 (3):\penalty0 589--602, March 2017.
\newblock ISSN 0003-0007, 1520-0477.
\newblock \doi{10.1175/BAMS-D-15-00135.1}.
\newblock URL \url{https://journals.ametsoc.org/view/journals/bams/98/3/bams-d-15-00135.1.xml}.

\bibitem[Rasp et~al.(2018)Rasp, Pritchard, and Gentine]{rasp_deep_2018}
Stephan Rasp, Michael~S. Pritchard, and Pierre Gentine.
\newblock Deep learning to represent subgrid processes in climate models.
\newblock \emph{Proceedings of the National Academy of Sciences}, 115\penalty0 (39):\penalty0 9684--9689, September 2018.
\newblock \doi{10.1073/pnas.1810286115}.
\newblock URL \url{https://www.pnas.org/doi/10.1073/pnas.1810286115}.
\newblock Publisher: Proceedings of the National Academy of Sciences.

\bibitem[Sharma and Greenberg(2024)]{sharma_superdropnet_2024}
Shivani Sharma and David Greenberg.
\newblock {SuperdropNet}: a {Stable} and {Accurate} {Machine} {Learning} {Proxy} for {Droplet}-based {Cloud} {Microphysics}, February 2024.
\newblock URL \url{http://arxiv.org/abs/2402.18354}.
\newblock arXiv:2402.18354 [physics].

\bibitem[Filoche et~al.(2022)Filoche, Brajard, Charantonis, and Béréziat]{filoche_learning_2022}
Arthur Filoche, Julien Brajard, Anastase Charantonis, and Dominique Béréziat.
\newblock Learning {4DVAR} inversion directly from observations, November 2022.
\newblock URL \url{http://arxiv.org/abs/2211.09741}.
\newblock arXiv:2211.09741 [cs].

\bibitem[Wang et~al.(2024)Wang, Ren, Duan, Zhu, Li, Ni, Lu, and Yuan]{wang_four-dimensional_2024}
Wuxin Wang, Kaijun Ren, Boheng Duan, Junxing Zhu, Xiaoyong Li, Weicheng Ni, Jingze Lu, and Taikang Yuan.
\newblock A {Four}-{Dimensional} {Variational} {Constrained} {Neural} {Network}-{Based} {Data} {Assimilation} {Method}.
\newblock \emph{Journal of Advances in Modeling Earth Systems}, 16\penalty0 (1):\penalty0 e2023MS003687, 2024.
\newblock ISSN 1942-2466.
\newblock \doi{10.1029/2023MS003687}.
\newblock URL \url{https://onlinelibrary.wiley.com/doi/abs/10.1029/2023MS003687}.

\bibitem[Fisher et~al.(2011)Fisher, Trémolet, Auvinen, Tan, and Poli]{fisher_weak-constraint_2011}
Mike Fisher, Y.~Trémolet, H.~Auvinen, D.~Tan, and P.~Poli.
\newblock Weak-constraint and long window {4DVAR}, 2011.
\newblock Pages: 47 Publisher: ECMWF.

\bibitem[Evensen(2003)]{evensen_ensemble_2003}
Geir Evensen.
\newblock The {Ensemble} {Kalman} {Filter}: theoretical formulation and practical implementation.
\newblock \emph{Ocean Dynamics}, 53\penalty0 (4):\penalty0 343--367, November 2003.
\newblock ISSN 1616-7228.
\newblock \doi{10.1007/s10236-003-0036-9}.
\newblock URL \url{https://doi.org/10.1007/s10236-003-0036-9}.

\bibitem[Ronneberger et~al.(2015)Ronneberger, Fischer, and Brox]{ronneberger_u-net_2015}
Olaf Ronneberger, Philipp Fischer, and Thomas Brox.
\newblock U-{Net}: {Convolutional} {Networks} for {Biomedical} {Image} {Segmentation}, May 2015.
\newblock URL \url{http://arxiv.org/abs/1505.04597}.
\newblock arXiv:1505.04597 [cs] version: 1.

\bibitem[Isola et~al.(2018)Isola, Zhu, Zhou, and Efros]{isola_image--image_2018}
Phillip Isola, Jun-Yan Zhu, Tinghui Zhou, and Alexei~A. Efros.
\newblock Image-to-{Image} {Translation} with {Conditional} {Adversarial} {Networks}, November 2018.
\newblock URL \url{http://arxiv.org/abs/1611.07004}.
\newblock arXiv:1611.07004 [cs].

\bibitem[Gupta and Brandstetter(2023)]{gupta_towards_2023}
Jayesh~K. Gupta and Johannes Brandstetter.
\newblock Towards {Multi}-spatiotemporal-scale {Generalized} {PDE} {Modeling}.
\newblock \emph{Transactions on Machine Learning Research}, January 2023.
\newblock ISSN 2835-8856.
\newblock URL \url{https://openreview.net/forum?id=dPSTDbGtBY}.

\bibitem[Lippe et~al.(2023)Lippe, Veeling, Perdikaris, Turner, and Brandstetter]{lippe_pde-refiner_2023}
Phillip Lippe, Bastiaan~S. Veeling, Paris Perdikaris, Richard~E. Turner, and Johannes Brandstetter.
\newblock {PDE}-{Refiner}: {Achieving} {Accurate} {Long} {Rollouts} with {Neural} {PDE} {Solvers}, October 2023.
\newblock URL \url{http://arxiv.org/abs/2308.05732}.
\newblock arXiv:2308.05732 [cs].

\bibitem[Lorenz(1995)]{lorenz_edward_n_predictability_1996}
E.N. Lorenz.
\newblock Predictability: a problem partly solved.
\newblock \emph{Seminar on Predictability, 4-8 September 1995}, 1:\penalty0 1--18, 1995 1995.

\bibitem[Cintra et~al.(2016)Cintra, de~Campos~Velho, and Cocke]{cintra_tracking_2016}
Rosangela Cintra, Haroldo de~Campos~Velho, and Steven Cocke.
\newblock Tracking the model: {Data} assimilation by artificial neural network.
\newblock In \emph{2016 {International} {Joint} {Conference} on {Neural} {Networks} ({IJCNN})}, pages 403--410, July 2016.
\newblock \doi{10.1109/IJCNN.2016.7727227}.
\newblock URL \url{https://ieeexplore.ieee.org/document/7727227}.
\newblock ISSN: 2161-4407.

\bibitem[Frerix et~al.(2021)Frerix, Kochkov, Smith, Cremers, Brenner, and Hoyer]{frerix_variational_2021}
Thomas Frerix, Dmitrii Kochkov, Jamie~A. Smith, Daniel Cremers, Michael~P. Brenner, and Stephan Hoyer.
\newblock Variational {Data} {Assimilation} with a {Learned} {Inverse} {Observation} {Operator}, May 2021.
\newblock URL \url{http://arxiv.org/abs/2102.11192}.
\newblock arXiv:2102.11192 [physics].

\bibitem[Penny et~al.(2022)Penny, Smith, Chen, Platt, Lin, Goodliff, and Abarbanel]{penny_integrating_2022}
Stephen~G. Penny, Timothy~A. Smith, Tse-Chun Chen, Jason~A. Platt, Hsin-Yi Lin, Michael Goodliff, and Henry D.~I. Abarbanel.
\newblock Integrating {Recurrent} {Neural} {Networks} with {Data} {Assimilation} for {Scalable} {Data}-{Driven} {State} {Estimation}.
\newblock \emph{Journal of Advances in Modeling Earth Systems}, 14\penalty0 (3), March 2022.
\newblock ISSN 1942-2466, 1942-2466.
\newblock \doi{10.1029/2021MS002843}.
\newblock URL \url{http://arxiv.org/abs/2109.12269}.
\newblock arXiv:2109.12269 [physics].

\bibitem[De~Almeida et~al.(2022)De~Almeida, De~Campos~Velho, França, and Ebecken]{de_almeida_neural_2022}
Vinícius~Albuquerque De~Almeida, Haroldo~Fraga De~Campos~Velho, Gutemberg~Borges França, and Nelson Francisco~Favilla Ebecken.
\newblock Neural networks for data assimilation of surface and upper-air data in {Rio} de {Janeiro}, September 2022.
\newblock URL \url{https://gmd.copernicus.org/preprints/gmd-2022-50/}.

\bibitem[Fablet et~al.(2021)Fablet, Chapron, Drumetz, Mémin, Pannekoucke, and Rousseau]{fablet_learning_2021}
R.~Fablet, B.~Chapron, L.~Drumetz, E.~Mémin, O.~Pannekoucke, and F.~Rousseau.
\newblock Learning {Variational} {Data} {Assimilation} {Models} and {Solvers}.
\newblock \emph{Journal of Advances in Modeling Earth Systems}, 13\penalty0 (10):\penalty0 e2021MS002572, 2021.
\newblock ISSN 1942-2466.
\newblock \doi{10.1029/2021MS002572}.
\newblock URL \url{https://onlinelibrary.wiley.com/doi/abs/10.1029/2021MS002572}.

\bibitem[Ruckstuhl and Janjić(2018)]{ruckstuhl_parameter_2018}
Y.~M. Ruckstuhl and T.~Janjić.
\newblock Parameter and state estimation with ensemble {Kalman} filter based algorithms for convective-scale applications.
\newblock \emph{Quarterly Journal of the Royal Meteorological Society}, 144\penalty0 (712):\penalty0 826--841, 2018.
\newblock ISSN 1477-870X.
\newblock \doi{10.1002/qj.3257}.
\newblock URL \url{https://rmets.onlinelibrary.wiley.com/doi/abs/10.1002/qj.3257}.

\bibitem[Qu et~al.(2024)Qu, Bhouri, and Gentine]{qu_joint_2024}
Yongquan Qu, Mohamed~Aziz Bhouri, and Pierre Gentine.
\newblock Joint {Parameter} and {Parameterization} {Inference} with {Uncertainty} {Quantification} through {Differentiable} {Programming}, May 2024.
\newblock URL \url{http://arxiv.org/abs/2403.02215}.
\newblock arXiv:2403.02215 [nlin, physics:physics].

\bibitem[Cranmer et~al.(2020)Cranmer, Brehmer, and Louppe]{cranmer_frontier_2020}
Kyle Cranmer, Johann Brehmer, and Gilles Louppe.
\newblock The frontier of simulation-based inference.
\newblock \emph{Proceedings of the National Academy of Sciences}, 117\penalty0 (48):\penalty0 30055--30062, December 2020.
\newblock \doi{10.1073/pnas.1912789117}.
\newblock URL \url{https://www.pnas.org/doi/full/10.1073/pnas.1912789117}.
\newblock Publisher: Proceedings of the National Academy of Sciences.

\bibitem[Farchi et~al.(2021)Farchi, Laloyaux, Bonavita, and Bocquet]{farchi_using_2021}
Alban Farchi, Patrick Laloyaux, Massimo Bonavita, and Marc Bocquet.
\newblock Using machine learning to correct model error in data assimilation and forecast applications.
\newblock \emph{Quarterly Journal of the Royal Meteorological Society}, 147\penalty0 (739):\penalty0 3067--3084, 2021.
\newblock ISSN 1477-870X.
\newblock \doi{10.1002/qj.4116}.
\newblock URL \url{https://onlinelibrary.wiley.com/doi/abs/10.1002/qj.4116}.

\bibitem[Bocquet et~al.(2019)Bocquet, Brajard, Carrassi, and Bertino]{bocquet_data_2019}
Marc Bocquet, Julien Brajard, Alberto Carrassi, and Laurent Bertino.
\newblock Data assimilation as a learning tool to infer ordinary differential equation representations of dynamical models.
\newblock \emph{Nonlinear Processes in Geophysics}, 26\penalty0 (3):\penalty0 143--162, July 2019.
\newblock ISSN 1023-5809.
\newblock \doi{10.5194/npg-26-143-2019}.
\newblock URL \url{https://npg.copernicus.org/articles/26/143/2019/}.
\newblock Publisher: Copernicus GmbH.

\bibitem[Bocquet et~al.(2020)Bocquet, Brajard, Carrassi, and Bertino]{bocquet_bayesian_2020}
Marc Bocquet, Julien Brajard, Alberto Carrassi, and Laurent Bertino.
\newblock Bayesian inference of chaotic dynamics by merging data assimilation, machine learning and expectation-maximization.
\newblock \emph{Foundations of Data Science}, 2\penalty0 (1):\penalty0 55--80, 2020.
\newblock ISSN 2639-8001.
\newblock \doi{10.3934/fods.2020004}.
\newblock URL \url{http://arxiv.org/abs/2001.06270}.
\newblock arXiv: 2001.06270.

\bibitem[Brajard et~al.(2020{\natexlab{a}})Brajard, Carassi, Bocquet, and Bertino]{brajard_combining_2020-1}
Julien Brajard, Alberto Carassi, Marc Bocquet, and Laurent Bertino.
\newblock Combining data assimilation and machine learning to emulate a dynamical model from sparse and noisy observations: a case study with the {Lorenz} 96 model.
\newblock \emph{Journal of Computational Science}, 44:\penalty0 101171, July 2020{\natexlab{a}}.
\newblock ISSN 18777503.
\newblock \doi{10.1016/j.jocs.2020.101171}.
\newblock URL \url{http://arxiv.org/abs/2001.01520}.
\newblock arXiv: 2001.01520.

\bibitem[Brajard et~al.(2020{\natexlab{b}})Brajard, Carrassi, Bocquet, and Bertino]{brajard_combining_2020}
Julien Brajard, Alberto Carrassi, Marc Bocquet, and Laurent Bertino.
\newblock Combining data assimilation and machine learning to infer unresolved scale parametrisation.
\newblock \emph{arXiv:2009.04318 [physics, stat]}, December 2020{\natexlab{b}}.
\newblock URL \url{http://arxiv.org/abs/2009.04318}.
\newblock arXiv: 2009.04318.

\end{thebibliography}

\end{document}